\documentclass[sigconf,review=false]{acmart}

\AtBeginDocument{%
  \providecommand\BibTeX{{%
    \normalfont B\kern-0.5em{\scshape i\kern-0.25em b}\kern-0.8em\TeX}}}

\setcopyright{acmlicensed}
\copyrightyear{2018}
\acmYear{2018}
\acmDOI{XXXXXXX.XXXXXXX}

\acmConference[ASPLOS '24]{$5^{th}$ International Workshop on Cloud Intelligence/AIOps }{April 27th, 2024}{San Diego, CA}
\usepackage{algorithm2e}
\usepackage{float}

\begin{document}

\title{Anomaly Detection for Incident Response at Scale}

\author{Hanzhang Wang}
\affiliation{%
  \institution{Walmart Global Tech}
  \streetaddress{860 W California Ave}
  \city{Sunnyvale}
  \state{California}
  \country{USA}
  \postcode{94086}}
\email{hanzhang.wang@walmart.com}

\author{Gowtham Kumar Tangirala}
\affiliation{%
  \institution{Walmart Global Tech}
  \streetaddress{221 River Street}
  \city{Hoboken}
  \state{New Jersey}
  \country{USA}
  \postcode{07030}}
\email{Gowthamkumar.Tangira@walmart.com}

\author{Gilkara Pranav Naidu}
\affiliation{%
  \institution{Walmart Global Tech}
  \streetaddress{15 John Street}
  \city{Toronto}
  \state{Ontario}
  \country{Canada}
  \postcode{M5V 3G6}}
\email{Gilkarapranav.Naidu@walmart.com}

\author{Charles Mayville}
\affiliation{%
  \institution{Walmart Global Tech}
  \streetaddress{860 W California Ave}
  \city{Sunnyvale}
  \state{California}
  \country{USA}
  \postcode{94086}}
\email{Charles.Mayville@walmart.com}

\author{Arighna Roy}
\affiliation{%
  \institution{Walmart Global Tech}
  \streetaddress{221 River Street}
  \city{Hoboken}
  \state{New Jersey}
  \country{USA}
  \postcode{07030}}
\email{Arighna.Roy@walmart.com}

\author{Joanne Sun}
\affiliation{%
  \institution{Walmart Global Tech}
  \streetaddress{860 W California Ave}
  \city{Sunnyvale}
  \state{California}
  \country{USA}
  \postcode{94086}}
\email{Joanne.Sun@walmart.com}

\author{Ramesh Babu Mandava}
\affiliation{%
  \institution{Walmart Global Tech}
  \streetaddress{860 W California Ave}
  \city{Sunnyvale}
  \state{California}
  \country{USA}
  \postcode{94086}}
\email{Rameshbabu.Mandava@walmart.com}

\renewcommand{\shortauthors}{H. Wang, et al.}


\begin{abstract}
    We present a machine learning-based anomaly detection product, AI Detect and Respond (AIDR), that monitors Walmart's business and system health in real-time. During the validation over 3 months, the product served predictions from over 3000 models to more than 25 application, platform, and operation teams, covering 63\% of major incidents and reducing the mean-time-to-detect (MTTD) by more than 7 minutes. Unlike previous anomaly detection methods, our solution leverages statistical, ML and deep learning models while continuing to incorporate rule-based static thresholds to incorporate domain-specific knowledge.
    
    Both univariate and multivariate ML models are deployed and maintained through distributed services for scalability and high availability. AIDR has a feedback loop that assesses model quality with a combination of drift detection algorithms and customer feedback. It also offers self-onboarding capabilities and customizability.
    
    AIDR has achieved success with various internal teams with lower time to detection and fewer false positives than previous methods. As we move forward, we aim to expand incident coverage and prevention, reduce noise, and integrate further with root cause recommendation (RCR) to enable an end-to-end AIDR experience.
\end{abstract}





\maketitle

\section{Introduction}
In this paper, we present our work in developing a real-time anomaly detection product for Walmart. In such a complex organization, there are multiple dedicated teams focused on tracking crucial system health metrics in real-time to ensure smooth day-to-day operations. Any single failure can potentially trigger a domino effect and impact multiple dependent systems, cascading into a major incident that directly affects Walmart's revenue stream. Therefore, it is imperative to detect and mitigate any such issue as soon as possible. Thus, anomaly detection (AD) in real-time holds great significance within Walmart's operational landscape as it plays an important role in maintaining a seamless customer experience.

Traditional real-time anomaly detection efforts were driven primarily through rule-based alerting and eyeball monitoring. These approaches are time consuming, incomplete in their coverage, and agnostic to interdependent data patterns. In order to achieve high recall, rule-based alerting systems often increase alert volume in order to increase coverage, resulting in very low precision. This causes alert fatigue and reduces the efficacy. In contrast, ML-based solutions, when configured and trained carefully, offer improved accuracy and reduced noise. However, models are time consuming to maintain and hard to scale. Additionally, high knowledge barriers often limit the practicality of ML-based solutions to domain engineers with little ML know-how, risking wasted time and reduced accuracy. 

To address these issues, our anomaly detection solution, AIDR, is designed with the following key offerings:

\begin{enumerate}
    \item \textbf{Easy-to-customize and Adaptive Solution Flow}: AIDR interfaces are built with an emphasis on user-friendly design that offers a high degree of customizability for building AD models while minimizing required ML knowledge. It also allows customers to integrate their domain-specific knowledge and constraints when configuring alerts.
    
    \item \textbf{API-Driven Design}: AIDR operates on a cloud infrastructure, enabling us to deliver a practical, scalable, self-service machine learning-based alert system that is capable of delivering accurate, smart, and adaptable alerts to any team that requires them. As a self-contained, API-driven, cloud-native end-to-end MLOps implementation, it scales efficiently and eliminates the necessity for any ad-hoc scripting.
    
    \item \textbf{Self-Healing Model Life-cycle with Stability}: AIDR includes a self-healing model monitoring module with automatic drift detection and feedback monitoring features. This significantly reduces the need for manual intervention to maintain the model life-cycle, thereby rendering the system fully automatic, resulting in a reliable model life-cycle management. 
\end{enumerate}
 
In this paper, we explore various aspects of AIDR in depth. In Section \ref{system engineering}, we provide a high-level engineering architecture overview of AIDR. Section \ref{methodology} focuses on the ML algorithms that underpin the solution. In Section \ref{system validation}, we discuss its adoption within our organization and outline its impact on various internal teams. We provide a conclusion in Section \ref{conclusion} and lay out plans for future development.

\section{System Engineering}\label{system engineering}

\begin{figure*}[h]
  \centering
  \includegraphics[width=\linewidth]{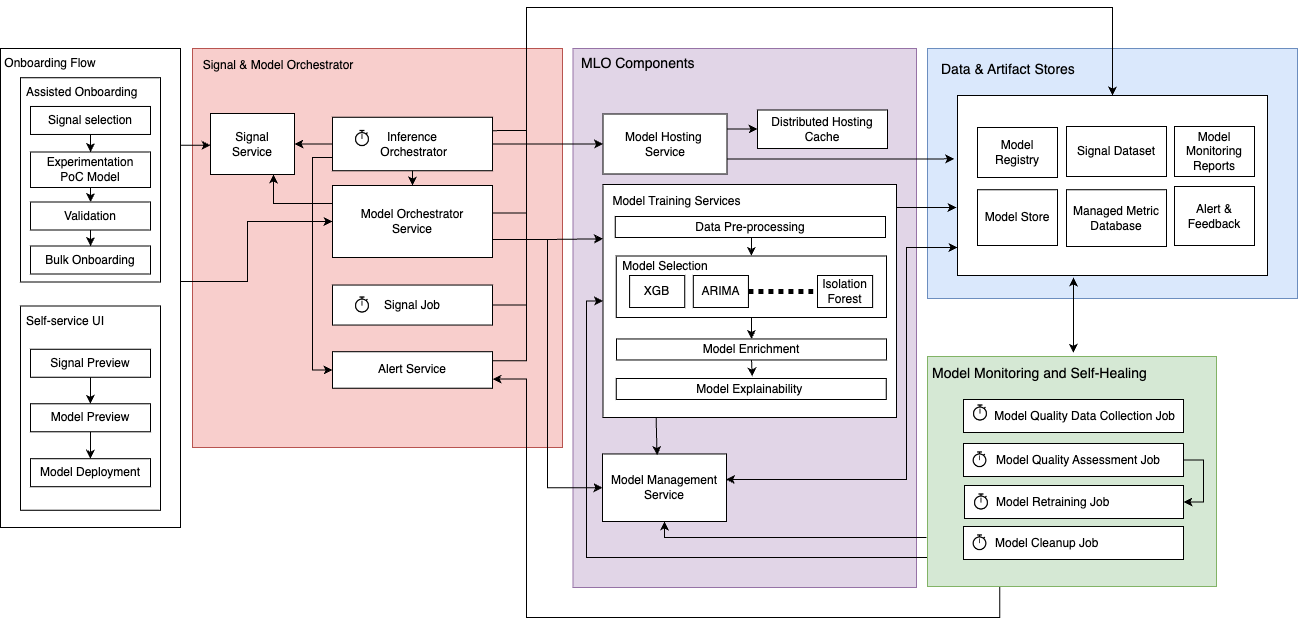}
  \caption{Anomaly Detection Platform Architecture}
  \label{arch-overview}
\end{figure*}

\subsection{Architecture Overview}\label{architecture overview}

As shown in Figure \ref{arch-overview}, the high-level architecture of this system is comprised of four main subsystems: Signal \& Model Orchestrator (SMO), Machine Learning Orchestrator (MLO), Data Store, and the Model Monitoring / Self-Healing Component. The SMO provides APIs to the onboarding UI which allows customers and data scientists to register signals and train, register, and deploy models. SMO dispatches model training, model preview and model management requests to the relevant MLO components. At every minute, SMO scans the model registry and fetches real-time metrics from the Prometheus-based Managed Metric Database (MMD) for each model's signals. Then, it calls Model Hosting Service to trigger predictions. The prediction scores and anomaly metrics are published back to MMD where they can be read by our own interfaces and customer tools. SMO also sends anomaly alerts as interactive Slack messages, allowing customers to mark those individual alerts as true or false positives, and stores that feedback data for future retraining. 

The Model Monitoring and Self-Healing Components empower our ML life-cycle. They detect potential data drift by taking into account several statistical measures (discussed in Section \ref{model monitoring}) via daily job. Once they identify the culprit models, they generate slack notifications to relevant stakeholders with a preview of suggested improvements to the model. If the user consents to the change, it seamlessly updates the model, which is utilized for real-time predictions immediately. This system allows us to preserve prediction quality and minimize manual intervention by both customers and data scientist during the models' life-cycles. 


The Data Store Component is responsible for storing all the ML artifacts such as signal data, serialized models, model registry, alert data and feedback data in a secure and organized manner. 

Overall, the SMO, MLO,  Model Monitoring and Self-Healing Components, and Data and Artifact Store work in tandem to ensure that the system is always learning and improving. With this architecture in place, AIDR can provide accurate and up-to-date predictions, making it an indispensable tool for site and platform reliability operations.

We have two routes to onboard customers to AIDR that takes advantage of the above-mentioned components. First, we provide a bespoke bulk onboarding strategy. Our data scientists collaborate with customers to determine the best selection of signals for their specific use case. They then build proof-of-concept (PoC) models for an initial validation before concurrently training and deploying hundreds of models. Each of these steps is executed by leveraging the API interface offered by SMO. This process typically takes about one month end-to-end for each customer. 

\begin{figure}
  \centering
  \includegraphics[width=\linewidth]{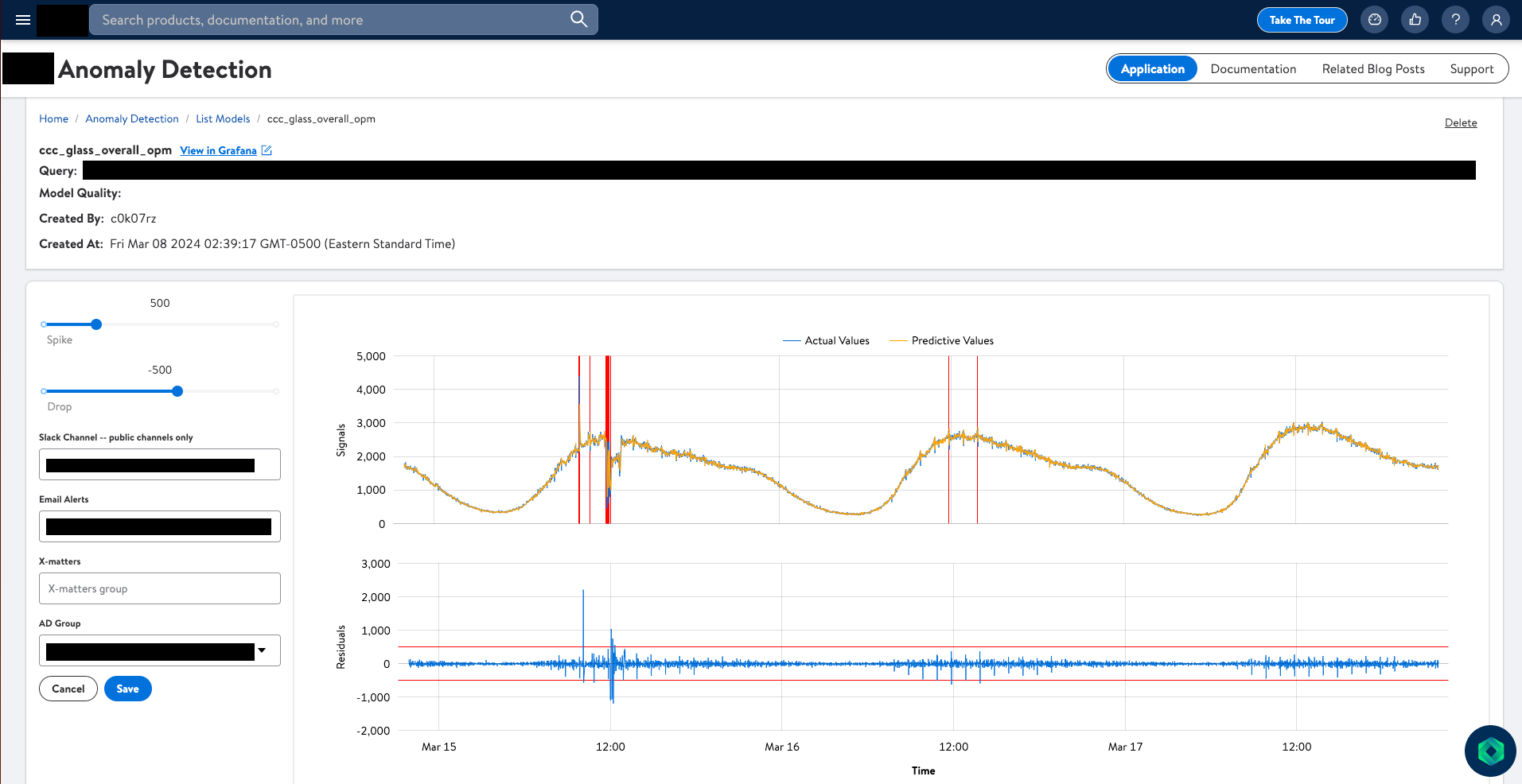}
  \caption{Anomaly Detection UI Univariate Model Onboarding}
  \label{dxio-onboarding}
\end{figure}

Additionally, we also offer a user-friendly self-onboarding UI portal where users with little to no ML knowledge can  visualize signals and train, preview, adjust, and deploy anomaly detection models in minutes with a few clicks of a button. They can independently maintain and manage these models and adjust parameters such as spike and drop thresholds in the univariate (UV) flow as in Figure \ref{dxio-onboarding}, or model sensitivity and hold variables in the multivariate (MV) flow.

\subsection{Signal \& Model Orchestrator} \label{ml orchestrator}

The SMO includes a variety of services including the Signal Service, Model Orchestrator Service, Inference Orchestrator, Signal Job, and Alert Service. These are designed to facilitate the onboarding UI for effective signal and model management. The SMO services are also instrumental in generating predictions and alerts, which are subsequently exposed to the MMD in the form of Prometheus metrics.


The first step of model onboarding involves signal registration, achieved through a Create Retrieve Update and Delete (CRUD) API interface. During signal registration, the specific signal name and the associated PromQL query are recorded. Once a signal is registered, 21 days' worth of time series data is collected from the MMD (the maximum held by MMD) and saved as a signal dataset in the Data Store. The daily Signal Job ensures this data remains current with every new run. Subsequent steps usually entail model training and registration, utilizing the underlying MLO interface for model training. Following this, the model is associated with the appropriate communication channel for publishing alert events. These requirements are met using the API contracts offered by the Model Orchestrator Service.


The Inference Orchestrator performs inference for all registered models on a minute-by-minute basis through the MLO components. It subsequently pushes alert events through the Alert Service upon the detection of an anomaly. It also exposes the corresponding anomaly metric and the predicted signal values into the MMD. The integration of these metrics into MMD allows for effective visualization of the original and predicted signals and associated anomalies on the user dashboard we provide.



To accommodate the increasing number of models and signals, the Inference Orchestrator implements a distributed load sharing approach which is horizontally scalable. This is achieved by utilizing a host discovery tool and Raft consensus protocol to select a leader host that orchestrates the load sharing process. Upon the creation of a new pod, the discovery service propagates the creation event to all inference service pods, thereby rendering the new pod discoverable. Leveraging the Raft consensus algorithm \cite{ongaro2014search}, a leader pod is subsequently elected within a few seconds to dispatch inference responsibilities to other stateless workers, thereby ensuring fault tolerance. The leader pod distributes the load to all non-leader pods using low latency gRPC. The non-leader pods perform prediction requests through MLO components, assigning prediction duties to them to prevent the occurrence of CPU spikes on the leader. Since the leader only distributes responsibilities, it never itself becomes a bottleneck.

Lastly, we make use of the Alert Service which provides a CRUD interface to publish alerts through Slack notifications. This allows the users to accomplish activities such as deleting or snoozing alerts seamlessly through buttons provided on the Slack notification. The alert data is also used to re-calibrate models using the Model Monitoring and Self-Healing components.

\subsection{MLO Components} \label{mlops}
The MLO components serve as building blocks to provide an interface for typical ML system requirements such as training, online inference and model configuration. All interactions with these components take place through the use of a versatile API-driven interface which ensures that a wide range of ML requirements can be fulfilled. Additionally, these fundamental services are designed to be stateless in nature, allowing the system to scale horizontally. 


\subsubsection{Model Hosting Service}
The Hosting service surfaces prediction APIs that provide the system online inference capabilities. It lets us support more than 4000 model inferences per minute while  accommodating various model types that may use entirely different ML stacks. 

The primary challenges lie in optimizing the model inference speed and scalability. The inference speed of the worker is inversely correlated to the scaling cost of the entire system. Higher inference time per worker implies more concurrent workers are required to perform prediction within a one-minute time period. This directly translates into higher scaling costs as we onboard additional models into our system. In order to optimize latency and scaling cost we make use of a distributed hosting cache and employ a multi-tiered load-balancing approach.

The distributed model cache stores serialized models in memory for swift retrieval prior to prediction. In case the model is not already present in cache, it is fetched from the Model Store and inserted into the cache for future use. This approach significantly improves prediction speed and ensures limitless scalability by decoupling model store from prediction, making the service stateless. 


Additionally, each hosting pod consists of several light-weight asynchronous workers. Leveraging the power of Kubernetes, we provide load-balancing at the worker, pod, and region levels, enhancing the resilience and fault tolerance of our system, resulting in a highly optimized inference service. In order to ensure this resiliency is maintained throughout the life of the system, releases undergo a thorough series of chaos testing.

\subsubsection{Model Management Service}
The Model Management Service (MMS) provides a CRUD API interface to interact with the Model Registry and Model Configuration. The Model Registry serves as a record of all onboarded models, their current versions, and metadata such as model type, training parameters (including holding and smoothing windows), and static thresholds. The SMO utilizes the model registry to identify all models for which it needs to periodically deliver and publish inferences. The Model Configuration---stored alongside each model in the Model Store---serves as a record of configurations specific to the model type, including parameters like minimum training length, minimum prediction step, and other model-specific parameters. These configurations are used by the SMO to prepare the appropriate payload for real-time inferences. They are also used by the Model Training Service to set the default values of optional training parameters.  

\subsubsection{Model Training Service}
Model Training Service provides APIs that initiate training, a process that creates versioned model artifacts that are able to be used in inference. The API provides the capability to execute both model preview and training. While generating the model preview, we make use of a smaller subset of the dataset to fast-train a temporary model which is intended to provide the user a sense of the effectiveness of the model in detecting anomalies. If the user is satisfied with the preview, they can proceed with model training. The model training module utilizes 21 days of historical data to train the model and create a model artifact that is stored in the Model Store. 

The training process starts with fetching the historical data from the Data Store and conducting data pre-processing. This is followed by Model Selection, in which the appropriate model training routine is identified and executed based on the model type specified in the API request. Once the model is trained, it goes through the Model Enrichment module for further refinement, to account for any smoothing or holding specified in the API request. We discuss this in detail in Section \ref{independent algorithms}. In the final step, multivariate models go through the Model Explainability module, which generates the Model Explainability object. This object provides insight into the underlying cause of the detected anomaly for root cause analysis. The final model object is packaged, serialized and stored in the Model Store. If a model is to be registered into production (specified in the API request), the model metadata is passed on to the Model Management Service which updates the model registry.

Model training is a complex and time-consuming process that requires significant computational resources. Maintaining stability and scalability while being cost-efficient becomes the top priority in our design.  We make use of Kubernetes-friendly serverless framework tailored to the specific requirements of model training, considering its infrequent usage and significant resource consumption. This serverless framework can automatically scale down to and up from zero, ensuring that we only utilize resources when there is a training request coming in.

\subsection{Model Monitoring and Self-Healing}\label{model monitoring}
We implemented a daily multiprocessing cron job designed to identify data and concept drift within each model. This was based on the deviation of the statistical distribution of signal data compared to the training data, along with the number of daily anomalies triggered. When a model is found to be faulty, a preview graph, visualizing models from both pre and post re-training, is dispatched to customers via Slack notifications. Model re-training automatically commences following customer approval or a timeout. The drift detection process, along with customer feedback, is used to refine the model parameters during the re-training of models. At present, data drift detection is grounded in five statistical methods: Kolmogorov-Smirnov test statistic, Population Stability Index (PSI), Kullback-Leibler divergence, Jensen-Shannon divergence, and Wasserstein distance.


\section{Methodology} \label{methodology}
Curating ML solutions for users with unique challenges requires a wide variety of models and model-enriching techniques. It also involves numerous iterations of model building. In this section, we lay out the end-to-end process of model onboarding in the context of these requirements.  

\subsection{Independent Algorithms} \label{independent algorithms}
AIDR is an amalgamation of ML predictive models, statistical estimators, explainable AI algorithms, and rule-based filtering. Our core ML model inventory accommodates a diverse set of algorithms that range from regression-based statistical models to advanced deep learning models. While deep learning models demonstrate superior predictive performance, they often suffer from performance bottlenecks in production and require additional attention \cite{thompson2020computational}. Further, less complicated solutions are often good fits for fast user adoption.


\subsubsection{Core Predictive Algorithms:} At the core of the anomaly detection engine, we have ARIMA (Auto-Regressive Integrated Moving Average) \cite{Shumway2017}, XGBoost \cite{chen2016xgboost}, Isolation Forest (IF) \cite{liu2008isolation}, and Variational Autoencoder \cite{Kingma2013} models, in conjunction with Inter-Quartile Range and Empirical Quantiles to estimate the respective decision boundaries for anomaly classification. 


\subsubsection{Model Enhancement Techniques:} These add-ons are designed to assist the core algorithms to handle certain challenges:
\begin{enumerate}
        \item \textbf{Noise injection:} Due to the data availability constraints in our Data Store, our ML models encounter sampling bias on the training data, especially during the onboarding phase. We employ a noise injection strategy in the training data to prevent overfitting of the trained model and make it immune to the benign spikes/ drops during inference \cite{9263602}.
        
        \item \textbf{Hold / Smoothing:} We further complement the predictive models with exponential smoothing \cite{gardner1985exponential} and hold logic on the anomaly scores to make the alerts resistant to temporary spikes/drops. Unlike intrusion detection, AIOps for failure management has a higher tolerance for data behavior change \cite{10.1145/3483424} \cite{7081762}. A detection alert is held back till the number of anomalous time points encountered within a given lookback window exceeds a certain limit. 
    
        \item \textbf{Rule-Based filter:} To further reduce the false positive rate, we implemented a rule-based filter which acts as a safety net for extreme cases through several static thresholds \cite{electronics9061017} \cite{5061947}. These thresholds can act as a cheap way to incorporate domain knowledge in the model. These are often set to be the Service-Level Objectives (SLO) of customers.
        
        \item \textbf{Explainable AI:} Beyond the predictive accuracy, we extend the responsibility of our AIML solution through model explainability. We employ a set of SHAP (SHapley Additive exPlanations) based explainers that act as surrogate models on top of the core predictive models. When an anomaly is detected, they help identify the major contributing signals.
        
        \item \textbf{Seasonality:} If seasonality is observed in the input streaming signals, we use MEDIFF \cite{li2020anomaly} to extract the seasonality component. MEDIFF is a moving median smoothing-based seasonality extraction algorithm.
\end{enumerate}

Algorithm \ref{flow}  shows a schematic to demonstrate the workflow of the system that describes how an incoming request is processed through the pipeline. 

\RestyleAlgo{ruled} 
\begin{algorithm}[h]
    \caption{Prediction flow of anomaly detection on streaming data}\label{flow}
    \KwData{ A rolling window of length $n$ sliced from the input streaming data}
    \KwResult{ is\_anomaly = True | False}
    \eIf{\# of time points breaching static thresholds $\leq$ hold tolerance }{
    is\_anomaly $\gets$ rule based decision\;  
    }{
        \tcc{if applicable} 
        seasonal\_component $\gets$ MEDIFF(Data)\; 
        
        anomaly\_scores $\gets$ core model prediction score(Data, seasonal\_component)\;
        
        anomaly\_counts $\gets$ exponential\_smoothing($anomaly\_scores$)\;
        
        is\_anomaly $\gets$ apply\_hold\_tolerance(anomaly\_counts)
    }
\end{algorithm}

\subsection{Model Onboarding Life-cycle}
A quick model onboarding process comes with the potential cost of surgical changes during future iterations. To handle that, we wrap each algorithm into self-contained capsules that can easily be plugged into ensemble solutions. It makes our initial drafts adaptive to the iterative requirement changes of the customers.
\begin{table}
    \caption{Comparative Analysis of Univariate Models for a Labeled Dataset}
    \label{uni}
        \begin{center}
            \begin{tabular}{|  c  |c  |c  |c  |}
                \hline
                Models & Precision & Recall & Balanced Accuracy\\
                \hline
                ARIMA & 0.98 & 0.53 & 0.76 \\
                \hline
                ARIMA+MEDIFF& 0.97 & 0.70 & 0.85 \\
                \hline
                XGBoost& 0.10 & 0.96 & 0.86 \\
                \hline
            \end{tabular}
        \end{center}
\end{table} 


As described in Figure \ref{uni}, we provide a self-onboarding process of models through an in-house UI supported by the SMO APIs. The plug-and-play model building interface, complemented with default configurations, provides an AutoML solution to customers with minimal ML know-how. Tier-1 customers require bulk onboarding and are supported by data scientists. Some of the models onboarded in bulk suffer from concept drift and turn into quiet and noisy models. A self-healing solution is designed to perform drift detection and autonomous re-training, as discussed in Section \ref{model monitoring}.


Our autonomous model maintenance system helps data scientists focus on new interesting use cases with unique problem statements. Our configuration-driven model-building pipeline makes it easy for data scientists to quickly iterate through the potential solutions during the ideation phase. Our centralized Artifact Store reduces the experimentation time to find a problem-solution fit while building PoC models. The non-intrusive nature of the algorithm plug-ins makes quick prototyping possible without risking the possibility of incremental efforts in the future. 


\subsection{Experimental Results}

\begin{table}
    \caption{Comparative Analysis of Balanced Accuracy for Multivariate Models}
    \label{mv}
        \begin{center}
            \begin{tabular}{|  c  |c  |c  |c  |c  |}
                \hline
                Dataset& IF & IF        & IF                    & Autoencoder \\
                       &    & (hold)    & (hold \& smoothing)   &             \\ 
                \hline
                DS 1& 0.70 & 0.74 & 0.75 & 0.81 \\
                \hline
                DS 2& 0.59 & 0.66 & 0.67 & 0.75 \\
                \hline
                DS 3& 0.64 & 0.70 & 0.71 & 0.72 \\
                \hline
                DS 4& 0.65 & 0.70 & 0.71 & 0.81 \\
                \hline
            \end{tabular}
        \end{center}
\end{table}


\begin{table}
    \caption{AIDR Noise Reduction and Success}
    \label{success-table}
    \begin{center}
        \begin{tabular}{ | c | c | c | c | }
            \hline
            Customer  & Noise & Incidents & Incidents \\
            ~ & Reduction & Detected & Prevented \\
            \hline
            Pricing \& Delivery & 94\%  & 100\% (4/4) & 2 \\
            \hline
            Payments & 90\% & 100\% (3/3) & 2 \\
            \hline
            Networks & 99\% & 100\% (3/3) & 2 \\
            \hline
            Operations & 91\% & 56\% (27/48)  & 1 \\                
            \hline
        \end{tabular}
    \end{center}
\end{table}

Table \ref{uni} shows the effectiveness of various supported model types for the UV flow through three evaluation metrics---precision, recall and balanced accuracy. These models were built for one of our point-of-sale traffic metrics and were validated over a period of 90 days. The models reported in Table \ref{uni} were arrived at by maximizing the balanced accuracy score. We chose to consider the balanced accuracy as our key performance metric due to the extreme imbalance observed in our datasets. This metric, unlike traditional accuracy, gives equal weight to both false positives and false negatives, making it particularly effective for evaluating model performance on imbalanced data \cite{brodersen2010balanced}. 

While an XGBoost classifier yields a superior recall of 96\%, it severely suffers from a low precision of 10\%.  In contrast, a regression-based next-time-point-prediction model ARIMA yields an impressive precision of 98\% while it suffers from a low recall of 53\%. In practice, none of these extreme trade-offs are welcome for most of the user stories even though they yield a reasonable balanced accuracy. After combining MEDIFF with ARIMA, we see a significant improvement of recall (70\% from 53\%) for a negligible loss in precision and a higher balanced accuracy.  

Table \ref{mv} extends the comparative analysis for the MV flow to 4 datasets---DS1, DS2, DS3 and DS4---reporting the maximum balanced accuracy for each model type. DS1 corresponds to one of our E-commerce applications, DS2 to an internal database, DS3 to a critical application team, and DS4 to a critical group of operational infrastructure metrics. These models were validated over a period of 59 days.  

For MV use cases, our first choice is Isolation Forest. It has a low memory requirement and linear time complexity. The accuracy of a vanilla IF can be improved with the add-on decision-enriching techniques for all four datasets. An Autoencoder outperforms Isolation Forest for every dataset at the cost of computational cost. 

\section{System Validation}\label{system validation}

In this section, we present a few validation metrics and KPIs collected over the course of one business quarter that demonstrate the effectiveness of AIDR. 


During the validation for over 90 days, the anomaly detection platform served out predictions from over 3000 models, providing real-time anomaly detection to around 30 application, internal platforms, operations teams. AIDR is the default go-to solution for anomaly detection within the platform site reliability team. These models alerted early for 37 major incidents and triggered at least 12 code changes, backouts, and other interventions that prevented potential incidents. Table \ref{success-table} summarizes our high-level success around alert efficacy and noise reduction for 4 case study customers. 

Our AD models designed for pricing and delivery applications have a 100\% incident coverage rate and helped reduce MTTD by 7 minutes, with a 91\% noise reduction compared to non-AIDR alerts. In two cases, our models were the only source of alerts for upcoming issues on unhealthy applications and directly led to code changes. For AD models built to track payment transactions, our models have a 100\% incident coverage rate with an MTTD reduction of 14 minutes. Models here also worked to reduce mean-time-to-triage (MTTT). They were able to trace issues to specific payment authorizers and tender types in contrast to existing alerts, with an average of 16 minutes reduction in MTTT. 

For our internal network platform, we built models that monitor network connections to detect distributed-denial-of-service (DDoS) attacks. These models helped prevent 4 attacks from becoming incidents and detected another 4 earlier than alternative alerting. These models have a 100\% success rate in detecting DDoS attacks, with a 99\% noise reduction compared to non-AIDR alerts. For one of our database platforms, we built models to identify servers with bad, overloaded queries. 70\% of these alerts were found to be actionable. 

For the Operations team, we built AD models to cover key business signals (including E-commerce traffic and metrics tracking online user journey) and critical application and network health metrics. Within our organization, the Operations team serves as the first responder to any major incidents, with its scope spanning multiple global markets. Consequently, our model coverage for this team is relatively lower, leading to lower incident coverage numbers. We were able to cover 56\% of major incidents with a MTTD reduction of 7.2 minutes and prevent 1 incident, with a noise reduction of 91\% compared to non-AIDR alerts. While the noise reduction in this case is highly desirable, there is significant room for improvement in terms of incident coverage. 

\section{Conclusion}\label{conclusion}
In conclusion, our paper has presented a comprehensive, innovative, and scalable anomaly detection solution that harnesses statistical and machine learning models to deliver accurate, real-time alerts to production systems. The tech novelty of our solution lies in its ease of use for users with little to no ML know-how, its API-driven, scalable cloud-native design, its utilization of advanced algorithms bolstered by non-intrusive and algorithm plug-ins, and its implementation of a self-healing module for automated model life-cycle management. Its adoption and impact within our organization has been significant, providing our customers with an alternative to traditional means of alerting. We have identified some areas where significant improvement can be made, particularly with the Operations team. Going forward, we would like to incorporate our solution with a Root Cause Recommendation (RCR) system and generative AI empowered ChatOps to provide an end-to-end incident detection and triaging experience for our customers.
\bibliographystyle{ACM-Reference-Format}
\bibliography{sample-base}


\end{document}